\documentclass[10pt,twocolumn,letterpaper]{article}

\usepackage{cvpr}
\usepackage{times}
\usepackage{epsfig}
\usepackage{graphicx}
\usepackage{amsmath}
\usepackage{amssymb}

\usepackage[breaklinks=true,bookmarks=false]{hyperref}

\cvprfinalcopy 

\def\cvprPaperID{****} 

\usepackage{subfig}
\usepackage{wrapfig}
\usepackage{algorithm}
\usepackage{algpseudocode}
\usepackage{blindtext}
\usepackage{titling}
\algblockdefx{MRepeat}{EndRepeat}{\textbf{repeat}}{}
\algnotext{EndRepeat}
\newcommand{\myparagraph}[1]{\noindent\textbf{#1}}
\def \hfillx {\hspace*{-\textwidth} \hfill}

 \makeatletter
 \DeclareRobustCommand\onedot{\futurelet\@let@token\@onedot}
 \def\@onedot{\ifx\@let@token.\else.\null\fi\xspace}
  
 \def\ie{i.e\onedot}

 \def\etal{\textit{et~al\onedot}}

 \makeatother

\begin{document}

\title{
Task-Free Continual Learning}

\author{Rahaf Aljundi\thanks{Rahaf Aljundi and Klaas Kelchtermans contributed equally to this work and are listed in alphabetical order.}  \quad Klaas Kelchtermans$^{ 	\ast}$ \quad  Tinne Tuytelaars\\
KU Leuven, ESAT-PSI, Belgium \\
{\tt\small firstname.lastname@esat.kuleuven.be}}


\date{}
\maketitle


\begin{abstract}
Methods proposed in the literature towards continual deep learning 
typically operate in a task-based sequential learning setup.
A sequence of tasks is learned, one at a time, with all data of current task available but not of previous or future tasks.
Task boundaries and identities are known at all times.
This setup, however, is rarely encountered in practical applications. 
Therefore we investigate how to transform 
continual learning to an online setup.
We develop a system that keeps on learning over time in a streaming fashion,
with data distributions gradually changing and without the notion of separate tasks. 
To this end, we build on the work
on Memory Aware Synapses, and show how this method can be made online
by providing a protocol to decide i) when to update the importance weights, ii) which data to use to update them, and iii) how to accumulate the importance weights at each update step. 
Experimental results show the validity of the approach in the context of two applications: (self-)supervised learning of a face recognition model by watching soap series and learning a robot to avoid collisions. 
\end{abstract}

\section{Introduction}
In machine learning, one of the most basic paradigms 
is to clearly distinguish between a training and testing phase.
Once a model is trained and validated, it switches to a test mode: the model gets frozen and deployed for inference on 
previously unseen data, without ever making changes to the model parameters again. 
This setup 
assumes a ‘static’ world, 
with no distribution shifts over time.
Further, it assumes a static task specification, so no new requirements in terms of output (e.g. novel category labels) or new tasks added over time. 
Such strong division between training and testing makes it easier to develop novel machine learning algorithms, 
yet is also very restrictive.

Inspired by biological systems, the field of incremental learning, 
also referred to as continual learning or lifelong learning~\cite{pentina15nips,silver2013lifelong,thrun1995lifelong},
aims at breaking this strong barrier between the training and testing phase. 
The goal is to develop 
algorithms that do not stop learning, 
but rather keep updating the model parameters over time. 
This holds the promise of a system that gradually accumulates knowledge, 
reaching increasingly better accuracy and better coverage as time passes.
However, it is practically not possible to store all previous data - be it due to storage constraints or for privacy reasons. 
Yet updating parameters based only on recent data introduces a bias towards that data and a phenomenon known as catastrophic interference, in other words degrading performance on older data
~\cite{french1999catastrophic,robins1995catastrophic}.


To make progress in this direction, several works have opted for a 
specific experimental setup, consisting of 
a sequence of distinct tasks, learned one after the other. 
Each time, only the data for the `current' task is available for training.
We refer to this as {\em task-based sequential learning}.
Training a shared model one task at a time has led to significant progress and new insights towards continual learning, such as different strategies for preserving the knowledge of previous tasks~\cite{li2016learning,kirkpatrick2016overcoming,aljundi2017memory,lee2017overcoming}. 
However, the methods developed in this specific
setup all too often depend on knowing the task boundaries. 
These boundaries indicate 
good moments to consolidate knowledge, namely after learning a task. 
Moreover, data can be shuffled within a task so as to guarantee i.i.d. data. 
In an online setting, on the other hand, data needs to be processed in a
streaming fashion
and data distributions might change gradually.


In this work, we aim at overcoming this requirement of hard task boundaries. 
In particular, we investigate how methods proposed for task-based sequential learning can be generalized to an online setting. 
This requires a protocol to determine when to consolidate knowledge. 
Moreover, we investigate the effect of keeping a small buffer with difficult 
samples. 
For the latter, we take inspiration from the field of reinforcement learning, namely experience replay \cite{dqn2014mnih}, although using much smaller replay buffers, unlike very recent work of Rolnick et al.~\cite{rolnick17continualexpreplay}.

Task-based sequential learning
has mostly been studied for image classification~\cite{li2016learning,aljundi2016expert,lee2017overcoming,zenke2017improved,rannen2017encoder}. 
Whenever the learner arrives at a new task, that is when learning on the previous task has converged, 
a standard procedure is to extend the output layer of the network with additional `heads' for each of the new task's categories. 
Instead, the output of our network is fixed. 
In our first application, learning to recognize faces, we cope with a varying number of categories by using an embedding rather than class predictions. 
In our second application, learning a lightweight robot to navigate without collisions, it is not the output labels that change over time but rather the environment. 
For both applications, data is processed in a streaming fashion. This is challenging, since the data is not i.i.d. causing samples within one batch to be unbalanced.

The contributions of this paper are as follows: i) We are the first to extend task-based sequential learning to free and unknown task boundaries in an online continual learning scenario; ii) We develop protocols to integrate an importance weight regularizer, MAS, in this online continual learning setting; iii) Our experiments on face recognition from TV series and on monocular collision avoidance prove the effectiveness of our 
method
in handling the distribution changes in the streaming data and stabilizing the online learning behaviour, resulting in knowledge accumulation rather than catastrophic interference and improved performance in all the test cases.

In the following we discuss related work
(section~\ref{sec:relatework}). We then describe our   online continual learning approach in section~\ref{sec:methos}. We validate our system in the experimental section~\ref{sec:experiments} and end with discussion and conclusion in section~\ref{sec:discussion}.

\section{Related Work}\label{sec:relatework}
\paragraph{Online Learning:}
Whereas in traditional offline learning, the entire training data has to be made available prior to learning the task, on the contrary online learning studies learning algorithms that learn to optimize predictive models over a stream of data instances sequentially. We refer to~\cite{bottou1998online,shalev2012online}
for surveys and overviews on the topic.

A first set of online learning algorithms consists of different techniques designed to learn a linear 
model~\cite{hazan2008adaptive,duchi2011adaptive,strehl2008online,hu2011online}. 
Online learning with kernels~\cite{kivinen2004online} extends this line of work to non-linear models, but the models remain shallow and their performance lags behind the modern deep neural networks.
Unfortunately, attempts towards online learning of neural networks suffer from issues like convergence, catastrophic interference and more. 
Some recent works include \cite{sahoo2017online,ramasamy2018online}, who both start from a small network and then adapt the capacity by adding more neurons as new samples arrive, while for online deep metric learning, \cite{li2018online} proposed a method based on stacking multiple metric layers. 

In terms of applications, the work of Pernici \etal~\cite{pernici17video, pernici2018memory} is similar to our first application scenario. They learn face identities in a self-supervised fashion via temporal consistency. They start from the VGG face detector and descriptor, and use a memory of detected faces. 
In contrast, we start from a much weaker pretrained model (not face-specific), and update the model parameters over time while they do not.

A joint problem in continual and online learning is catastrophic interference~\cite{mccloskey1989catastrophic,french1999catastrophic} which is the severe forgetting of previous samples 
when learning new ones. This phenomenon manifests itself at different scales: in online learning it happens while learning samples with different patterns than previous ones; in the traditional setting of continual learning it happens over a sequence of tasks.\\[-20pt]
\paragraph{Continual Learning:}
In~\cite{hsu2018re}, Hsu \etal~classify 
the studied scenarios for continual learning into incremental task learning, incremental domain learning and incremental class learning.
They argue that more attention should go to the last two - \ie to methods that do not require to know the task identity, since that is the case encountered in most practical scenarios. 

Yet as indicated before, most methods to date follow the task-based sequential learning setup. This includes various 
regularization-based methods, such as Elastic Weight Consolidation~\cite{kirkpatrick2016overcoming}, Synaptic Intelligence~\cite{zenke2017improved} and Memory Aware Synapses~\cite{aljundi2017memory}.
These methods estimate importance weights for each model parameter and penalize changes to parameters deemed important for previous tasks. We will discuss how to extend one of them to the online setting later. Note that, while Synaptic Intelligence computes the importance weights in an online fashion, it still waits until the end of a task before updating the losses, so like the other methods it depends on knowing the task boundaries. Incremental Moment Matching~\cite{lee2017overcoming} builds on similar ideas, yet stores different models for different tasks and merges them only at the very end. As such, it is unclear how this could be extended to an online, task-free setting. 
Also related is the work on Dynamically Expandable Networks~\cite{yoon2018lifelong}. They exploit the relatedness between the new task and previously learned tasks to determine which neurons can be reused and add new neurons to account for the new knowledge.

Next there are several data-driven methods such as Learning without Forgetting~\cite{li2016learning} or Encoder-based Lifelong Learning~\cite{rannen2017encoder}. With a separate knowledge distillation loss term for previous tasks, it's again unclear how they could be applied without knowing the task identity.

Other methods use an episodic memory,
such as iCARL (incremental Classifier and Representation Learning)~\cite{rebuffi2016icarl} 
and Memory Based Parameter Adaptation~\cite{sprechmann2018memory}.
A special mention here goes to
Gradient Episodic Memory for Continual Learning~\cite{lopez2017gradient}, as it moves a step forward towards the online setting: it assumes that the learner receives examples one by one but simplifies the scenario to locally i.i.d. drawn samples from a task distribution. Moreover, it still assumes that a task identifier is given. Like the buffer we use, they use an episodic memory for each task consisting of recently seen examples.
A buffer from which recent data can be reused for training is similar to the concept of a replay buffer often used in Deep Reinforcement Learning (DRL). 
However a crucial difference is that in both old and recent DRL works the replay buffer typically contains up to 1M samples corresponding to over 100 days of experience \cite{dqn2014mnih,rainbow17hessel}.
Here, we want to keep the algorithm more online by using a buffer of up to 100 samples only.

A common DRL technique, known as ``prioritized sweeping", is to sample experiences with a large error more often than others \cite{Moore1993}.
In a similar fashion we propose ``prioritized keeping" where a hard buffer drops the easy samples first rather than the oldest.

\section{Method}\label{sec:methos}
Our goal is to design a training method for task-free online continual learning. 
Task-based sequential learning methods assume that data comes in tasks, with tasks boundaries identified, so the training procedure can be divided in consecutive phases. 
In between the training phases, when training has stabilized, the continual learning method updates its meta-knowledge on how to avoid forgetting previous tasks. 
However, in the case of online learning where data is streaming and the distribution is shifting gradually, it is unclear whether these methods can be applied and how.
%

After studying a couple of methods mentioned above, we identified Memory Aware Synapses (MAS)~\cite{aljundi2017memory} as the most promising method in this respect.
It enjoys the following favorable characteristics. 1) {\em Static storage requirement}: it only stores an importance weight for each parameter in the network avoiding an increase of memory consumption over time; 2) {\em Task agnostic}: it can be applied to any task and is not limited to classification. In particular, we can use it with an embedding as output, avoiding the need to add extra `heads' for new outputs over time; 3) {\em Fast}: it only needs one backward pass to update the importance weights. During training, the gradients of the imposed penalty are simply the change that occurs on each parameter weighted by its importance. Therefore, the penalty gradients can be added locally and do not need 
 a backpropagation step; 4) {\em top performance}: MAS shows superior performance to other importance weight regularizers~\cite{aljundi2017memory,hsu2018re}. 
In order to deploy MAS in an online continual learning scenario, we need to determine i) when to update the importance weights, ii) which data to use to update the importance weights, and iii) how to accumulate the importance weights at each update step.

We first introduce the considered online continual learning setup, then explain MAS and our training procedure under this setup.
  \begin{figure}[t]
\centering
 \vspace*{-0.7cm} 
\includegraphics[width=0.4\textwidth]{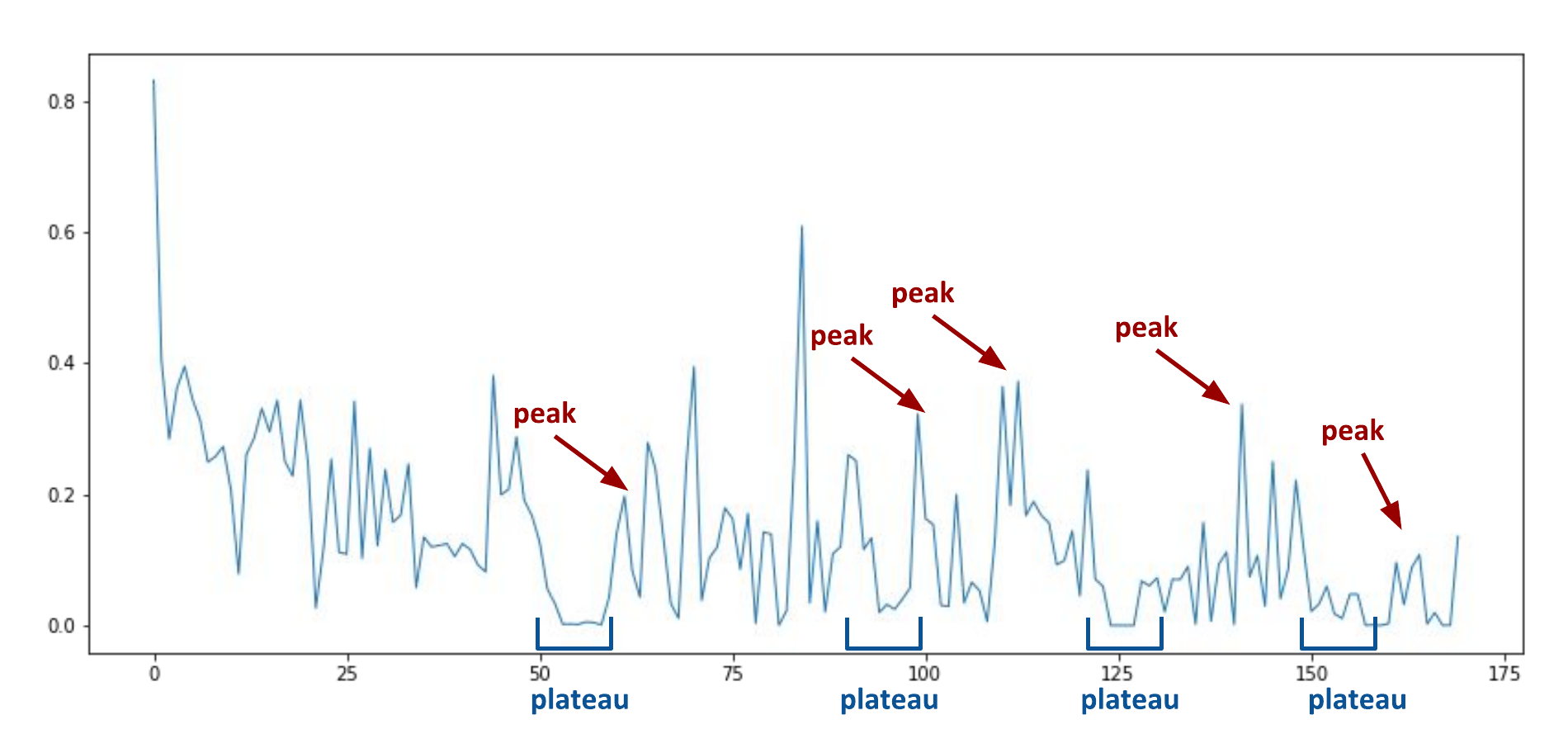}
  \caption{\footnotesize By detecting plateaus and peaks in the loss surface our method decides when to 
  update the importance weights, corresponding to the Big Bang Theory experiment, see section~\ref{sec:experiments:tvseries}; x-axis represents update steps}
  \label{fig:examples}
  \vspace*{-0.5cm} 
  \end{figure}
  
\myparagraph{Setup:}
We assume an infinite stream of data and a supervisory or self-supervisory signal that is generated based on few consecutive samples. At each time step $s$, the system receives a few consecutive samples along with their generated labels $\{x_k,y_k\}$ drawn non i.i.d from a current distribution $D_t$. Moreover, the distribution $D_t$ could itself experience sudden or gradual changes from $D_t$ to $D_{t+1}$ at any moment. The system is unaware of when these distribution changes are happening. The goal is to continually learn and update a function $F$ that minimizes the prediction errors on previously seen and future samples. In other words, it aims at continuously updating and accumulating knowledge. 
Given an input model with parameters $\theta$, the system at each time step reduces
the empirical risk based on the recently received samples and a small buffer $B$ composed of updated hard samples $X_B$. 
The learning objective of the online system is:
\begin{equation}
\underset{\theta}{min}~ \mathcal{L}( F(X;\theta),Y)+\mathcal{L}(F(X_{B};\theta),Y_{B})    
\end{equation}
Due to the strong non-i.i.d conditions and the very low number of samples used for the gradient step, the system is vulnerable to catastrophic interference between recent samples and previous samples and faces difficulty in accumulating the knowledge over time.

\myparagraph{Memory Aware Synapses (MAS)~\cite{aljundi2017memory}:}
In a traditional task-based sequential learning setting, MAS works as follows.
After each training phase (task), the method estimates an importance weight for each network parameter indicating the importance of this parameter to the previously learned task. To estimate the importance, MAS computes the sensitivity of the learned function to the parameters changes.
\begin{equation}\label{eq:1}
   F(x_k; \theta+ \delta) -F(x_k; \theta)\approx \sum_{i} g_{i}(x_k)  \delta_{i}
\end{equation}
\begin{equation}\label{eq:2}
\Omega_{i} =\frac{1}{N}\sum_{k=1}^N  \mid\mid g_{i}(x_k) \mid\mid
\end{equation}
where 
$\{x_k\}$ are the $N$ samples from the previous task, $\delta_i$ is a small change to model parameter $\theta_i$ and 
$g_{i}(x_k) = \frac{\partial{F(x_k)}}{\partial{\theta_{i}}}$. $\Omega_{i}$ is the importance weight of parameter $\theta_{i}$. When learning a new task, changes to important parameters are penalized:
\begin{equation}
L(\theta)=L_n(\theta)+\frac{\lambda}{2} \sum_{i} \Omega_{i}(\theta_{i}-\theta^*_{i})^2
\end{equation}
with $\theta^*$ the parameters values at the time of importance weight estimation, i.e. the optimal parameters for the previous task in the traditional sequential setup. $L_n(\theta)$ is the loss for the new task. After each task the newly estimated $\Omega_{i}$ are accumulated to the previous estimates.


         
  
\begin{algorithm}[t]
\caption{Online Continual Learning}\label{algo}
  \begin{algorithmic}[1]
  \footnotesize
    \State Input:$\delta_{\mu}, \delta_{\sigma}, \mathcal{N}$
    \State Initialize: $\mathcal{B}=\{\},\mathcal{W} =\{\},$
    \State $\Omega=\vec{0}, \mu^{old}_L=0, \sigma^{old}_L=0$, $\mathcal{P} =0$ 
 \MRepeat
    \State Receive K recent samples $X,Y$
    \For{$n$ in $\mathcal{N}$}
        \State $\mathcal{L}^{T}=\mathcal{L}_{\theta}(X,Y)+\mathcal{L}_{\theta}(X_{\mathcal{B}},Y_{\mathcal{B}})+\frac{\lambda}{2} \sum_{i} \Omega_{i}(\theta_{i}-\theta^*_{i})^2$
        \State $\theta\leftarrow$SGD($\theta,\mathcal{L}^{T})$
        \If{ $n=1$ }
            \State  $\mathcal{W} \leftarrow \text{update}(\mathcal{W}, \mathcal{L}(X,Y),\mathcal{L}(X_\mathcal{B},Y_\mathcal{B}))$
        \EndIf
    \EndFor
    \If{$\lnot \mathcal{P}$ $\land$ $\mu(\mathcal{W})<\delta_\mu$ $\land$ $\sigma(\mathcal{W})<\delta_\sigma$} 
         \State $\Omega \leftarrow \text{update}(\Omega,\theta,(X_\mathcal{B},Y_\mathcal{B}))$ 
          \State $\theta^*\leftarrow \theta$
         \State $\mu^{old}_L=\mu(\mathcal{W}), \sigma^{old}_L=\sigma(\mathcal{W})$
         \State $\mathcal{W}=\{\}$, $\mathcal{P}=1$
    \EndIf
    \If{$\mu(\mathcal{W})>\mu^{old}_L+\sigma^{old}_L$}
        \State $\mathcal{P}=0$
    \EndIf
    \State $(X_\mathcal{B},Y_\mathcal{B})\leftarrow\text{update}((X_{\mathcal{B}},Y_{\mathcal{B}}),(X,Y), \mathcal{L}_{\theta}(X,Y))$
  \EndRepeat
  \end{algorithmic}
\end{algorithm}

\myparagraph{When to update importance weights:}
In case of a task-based sequential learning setting where tasks have predefined boundaries, importance weights are updated after each task, when learning has converged. In the online case, the data is streaming without knowledge of a task's start or end (\ie when distribution shifts occur). So we need a mechanism to determine when to update the importance weights. For this, we look at the surface of the loss function. 

By observing the loss, we can derive some information about the data presented to the system.
When the loss 
decreases, this indicates that the model has learned some meaningful new knowledge from those seen samples. Yet the loss does not systematically decrease all the time. When new samples are received that are harder or contain different objects or input patterns than what was presented to the learner before, the loss may increase again. In those cases, the model has to update its knowledge, while minimally interfering with what has been learned previously. 

We can conclude that plateaus in the loss function indicate stable learning regimes, where the model is confidently predicting the current labels, see Figure~\ref{fig:examples}. Whenever the model is in such a stable area, it's a good time to consolidate the knowledge by updating the importance weights. This way, we identify the parameters that are important for the currently acquired knowledge. When learning new, ``different" samples the model will then be encouraged to preserve this knowledge.
This should allow the model to accumulate knowledge over time rather than replace previously learned bits of knowledge. 

\myparagraph{Detecting plateaus in the loss surface:}
To detect these plateaus in the loss surface, we use a sliding window over consecutive losses during training. We monitor the mean and the variance of the losses in this window and trigger an importance weight update whenever they are both lower than a given threshold. 
We do not keep re-estimating importance weights: we only re-check for plateaus in the loss surface after observing a peak. Peaks are detected when the window loss mean becomes higher than 85\% of a normal distribution estimated on the loss window of the previous plateau - that is when 
$\mu(L_{win})>\mu^{old}_L+\sigma^{old}_L$ where $\mu^{old}_L$ and $\sigma^{old}_L$
are the statistics of the previously detected plateau. This accounts for the continuous fluctuations in the loss function in the online learning and detects when significantly harder samples are observed.

\myparagraph{A small buffer with hard samples:} In a task-based sequential learning setup, importance weights are estimated on all the training data of the previous task. This is not an option for online learning, as storing all the previous data violates the condition of our setup. On the other hand, using only the most recent sequence of samples would lead to misleading estimates 
as 
these few consecutive samples might not be representative and hence do not capture the acquired knowledge correctly. 
To stabilize the online learning, we use a small buffer of hard samples that is updated at each learning step by keeping the samples with highest loss among the new samples and the current buffer. This is important as previous samples cannot be revisited and hence gives the system the advantage to re-process those hard samples and adjust its parameters towards better predictions in addition to getting a better estimate of the gradient step by averaging over the recent and hard samples. Moreover, the hard buffer represents a better estimate of the acquired knowledge than the few recent samples, hence allows for a better identification of importance weights. 

\myparagraph{Accumulating importance weights:} As we frequently update the importance weights, simply adding the new estimated importance values to the previous ones as suggested in MAS~\cite{aljundi2017memory} would lead to very high values and exploding gradients. Instead, we maintain a cumulative moving average of the estimated importance weights. Note, one could deploy a decaying factor that allows replacing old knowledge in the long term. However, in our experiments a cumulative moving average showed more stable results. 

After updating the importance weights, the model continues the learning process while penalizing changes to parameters that have been identified as important so far. As such our final learning objective is: 
\begin{equation}
\underset{\theta}{min} \,
\mathcal{L}(F(X;\theta),Y)+\mathcal{L}(F(X_B;\theta),Y_B)  +\frac{\lambda}{2} \sum_{i} \Omega_{i}(\theta_{i}-\theta^*_{i})^2 
\end{equation}
where $\theta^*$ are the parameters values at the last importance weight update step.  Algorithm~\ref{algo} summarizes the different steps of the proposed continual learning system.

\section{Experiments}\label{sec:experiments}
As a proof-of-concept, we validate our proposed method on a simple synthetic experiment. 
Later, we evaluate the method on two applications with either weak or self-supervision. 
First, we learn actor identities from watching soap series. 
The second application is robot navigation. 
In both cases, data is streaming and online continual learning is a key factor.

\begin{figure}[t]
\centering
\includegraphics[width=0.4\textwidth]{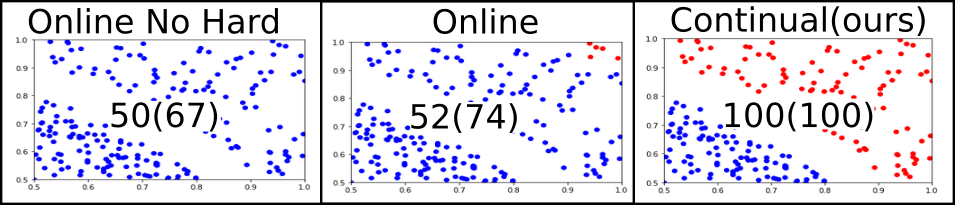}
\caption{\footnotesize Synthetic experiment: Predictions on first quadrant after training second quadrant. Test accuracy on first quadrant (total test accuracy on both quadrants) overlaid.}
\label{fig:synthetic}
\vspace*{-0.5cm} 
\end{figure}

\subsection{Synthetic Experiment}\label{sec:experiments:synthetic}
We constructed a binary classification problem with points in 4D in/out of the unit sphere. 
On a sequence of two tasks where each task corresponds to a quadrant, we test the performance of i) online without hard buffer, ii) online and iii) our full online continual method. 
Fig.\ref{fig:synthetic} depicts the predictions near the decision boundary in the first quadrant at the end of training on data in the second quadrant. 
The hard buffer results in better learning (higher total test accuracy), while the full method also avoids forgetting.

\subsection{Continual Learning by watching Soap Series}\label{sec:experiments:tvseries}
Here, we assume that an intelligent agent is watching episodes from a tv series and learns to differentiate between the faces of the different actors. The agent is equipped with a face detector module that is detecting faces online and a multi-object face tracker. In the case of weak supervision, we assume there is an annotator telling the agent whether two consecutive tracks are of the same identity or not. For the self-supervised case, we use the fact that if two faces are detected in the same image then their tracks must belong to two different actors.

\myparagraph{Setup:}
We start from an AlexNet~\cite{krizhevsky2014one} architecture with the convolutional layers pre-trained on ImageNet~\cite{krizhevsky2012imagenet} and the fully connected layers initialized randomly. The output layer is of size 100. Since the input consists of two tracks of two different identities, we use the triplet margin loss~\cite{balntas2016learning} which has been shown to work well in face recognition applications. This has the additional advantage that we don't need to know all the identities beforehand and new actors can be added as more episodes are watched. 

\myparagraph{Dataset:}
We use the actor labelling dataset from~\cite{aljundi2016actor}, specifically 6 episodes of The Big Bang Theory (BBT), 4 episodes of Breaking Bad (BB), and one episode of Mad Men (MM)\footnote{Unfortunately, there was an issue with the labels for the other episodes of Mad Men, which prevented us from using these.}. Note that for BB and MM, the episodes were further split into a total of 22 and 5 chunks, respectively.
For each episode we use the frames, detected faces and tracks along with track labels from~\cite{aljundi2016actor}. 
Tracks are processed in chronological order, imitating the setting where tracks are extracted in an online fashion while watching the tv series. As a result, the data is clearly non-i.i.d..

For the supervised setup, every tenth/fifth track is held out as test data in BBT/BB respectively as the latter has more tracks, 339 tracks BBT compared to 3941 BB. All the other tracks are used for training.
As we only have one episode for MM, we decided not to use it for the supervised setup.

For the self-supervised setup, BB turned out to be unsuitable, given that it is an actor centric series with a large majority of the scenes focusing on one actor. To still have results on two series, we do report also on MM in this case, in spite of it being only one episode.
Further, the original tracks provided by~\cite{aljundi2016actor} were quite short (an average of 8/22 faces per track in BBT/MM).  Since this is problematic for the self-supervised setting, we use a simple heuristic based on the distance between the faces embedding (based on AlexNet pretrained on ImageNet) to merge adjacent tracks belonging to the same actor.

\myparagraph{Training:}
Whenever two tracks are encountered belonging to different actors, a training step is performed using the detected faces (one face every 5 frames). If the two tracks contain more than 100 faces, a random sampling step is performed. We use a hard buffer size of 100 triplets and a fixed loss window size of 5. A few gradient steps are performed at each time step (2-3 for the supervised setting, 10-15 for the self-supervised one). We use SGD optimizer with a learning rate of $10^{-4}$. Hyperparameters were set based on the first BBT episode, please refer to the supplementary material for more details.

\myparagraph{Test:}
To test the accuracy of the trained model on recognizing the actors in the tv series, we use 5 templates of each actor selected from different episodes. 
We then compute the Euclidean distance of each test face to the templates, based on the learned representation, and assign the input face to the identity of the template that is closest. 
\begin{figure*}[t]

    \centering
    \subfloat[]{{\includegraphics[width=0.356\textwidth]{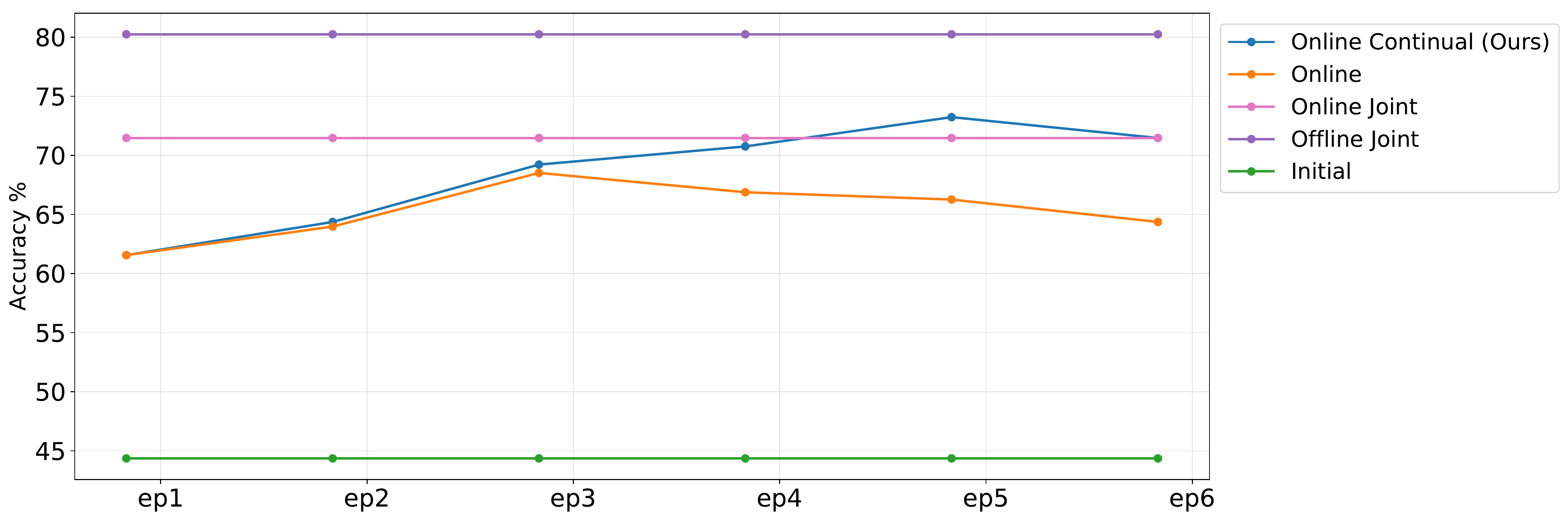} }}%
     \hfillx
    \subfloat[]{{\includegraphics[width=0.356\textwidth]{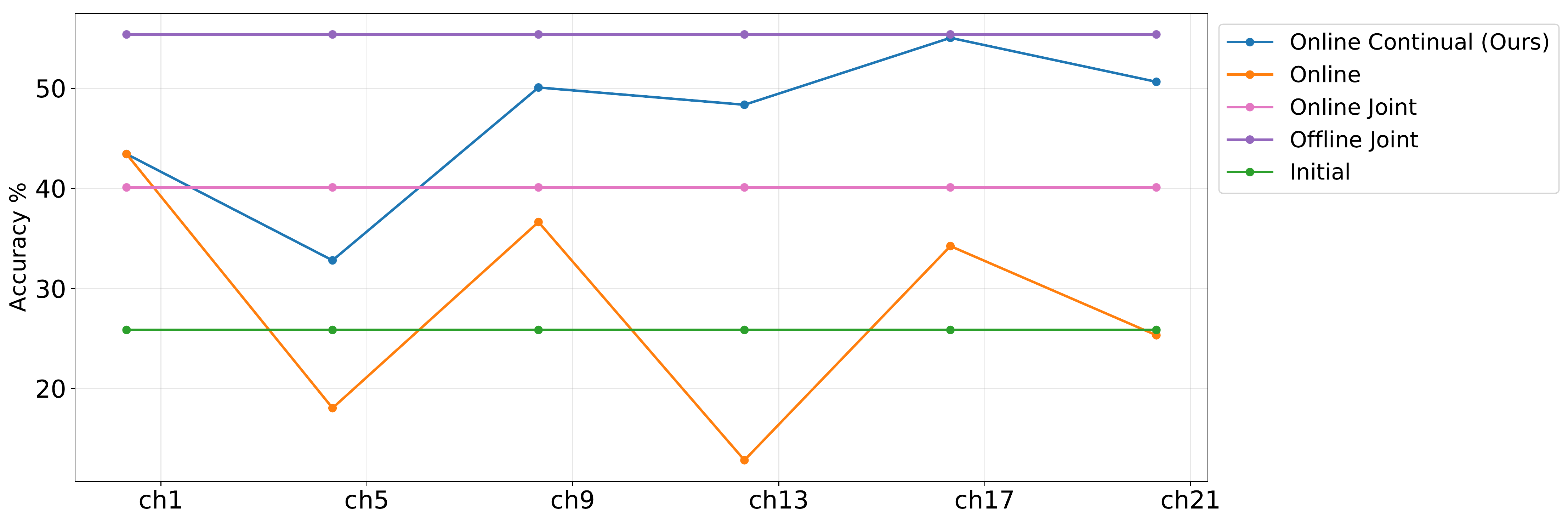} }}%
      \subfloat[]{{\includegraphics[width=0.356\textwidth]{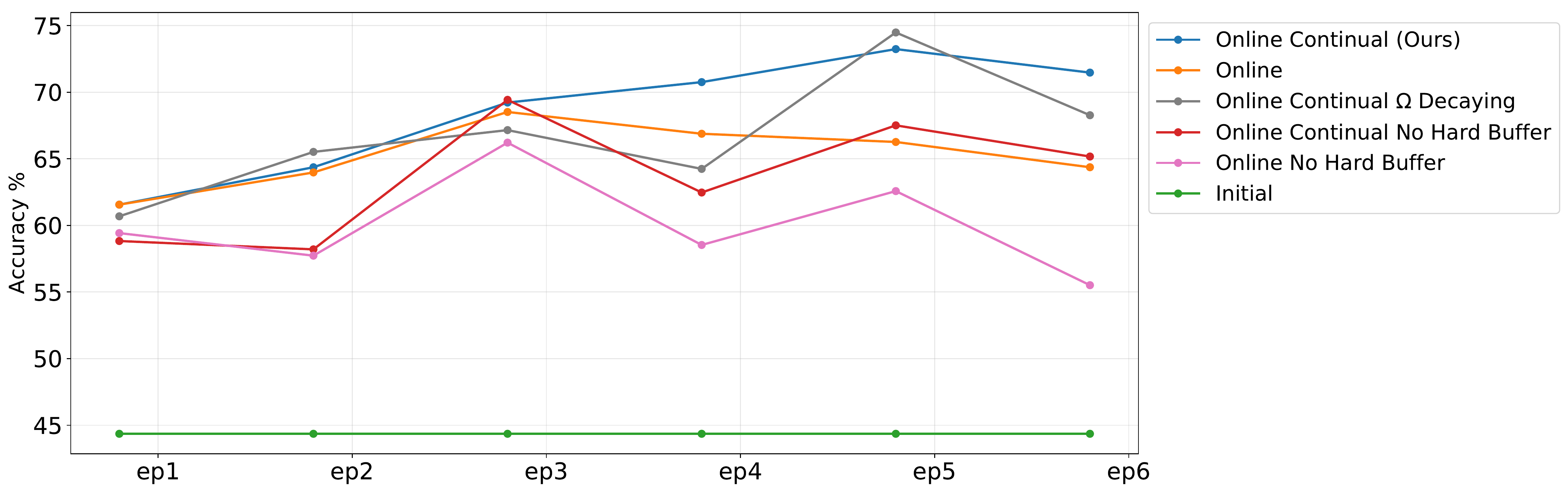} }}%
        \vspace*{-0.4cm}
         \caption{\footnotesize  Accuracy on test data at the end of each episode for Big Bang Theory (a) and after chunks 1,5,9,13,17 and 21 of Breaking Bad (b). (c) a study on the importance of the hard buffer and the cumulative $\Omega$ average versus a decaying $\Omega$, figure shows the test accuracy after each episode of BBT.}
         \label{fig:soapseries_super}%
\end{figure*}

\myparagraph{Baselines:}
To estimate the benefit of our system, {\em Online Continual}, we compare it against the following baselines:
\begin{enumerate}
    \item {\em Initial} : the pretrained model, \ie before training on any of the episodes. 
    \item {\em Online Baseline} : a model trained in the explained online setting but without the MAS importance weight regularizer.
    \vspace*{-0.2cm}
    \item {\em Online Joint Training} : a model trained online, again without MAS regularization, but with shuffled tracks across episodes to obtain i.i.d. drawn data.
    \item {\em Offline Joint Training} : a model that differs from Online Joint Training by going multiple epochs over the whole data. This stands as an upper bound.
    \vspace*{-0.4cm}
\end{enumerate}
\begin{figure*}%
\vspace*{-0.5cm}
    \centering
    \subfloat[]{{\includegraphics[width=0.39\textwidth]{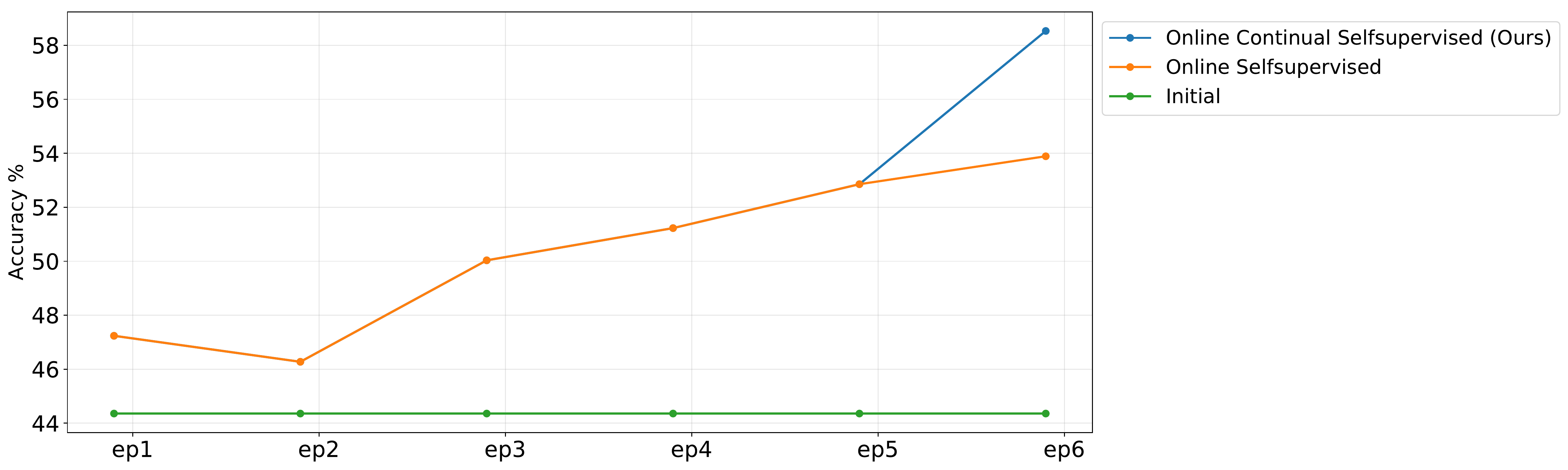} }}%
    \hfillx
    \subfloat[]{{\includegraphics[width=0.365\textwidth]{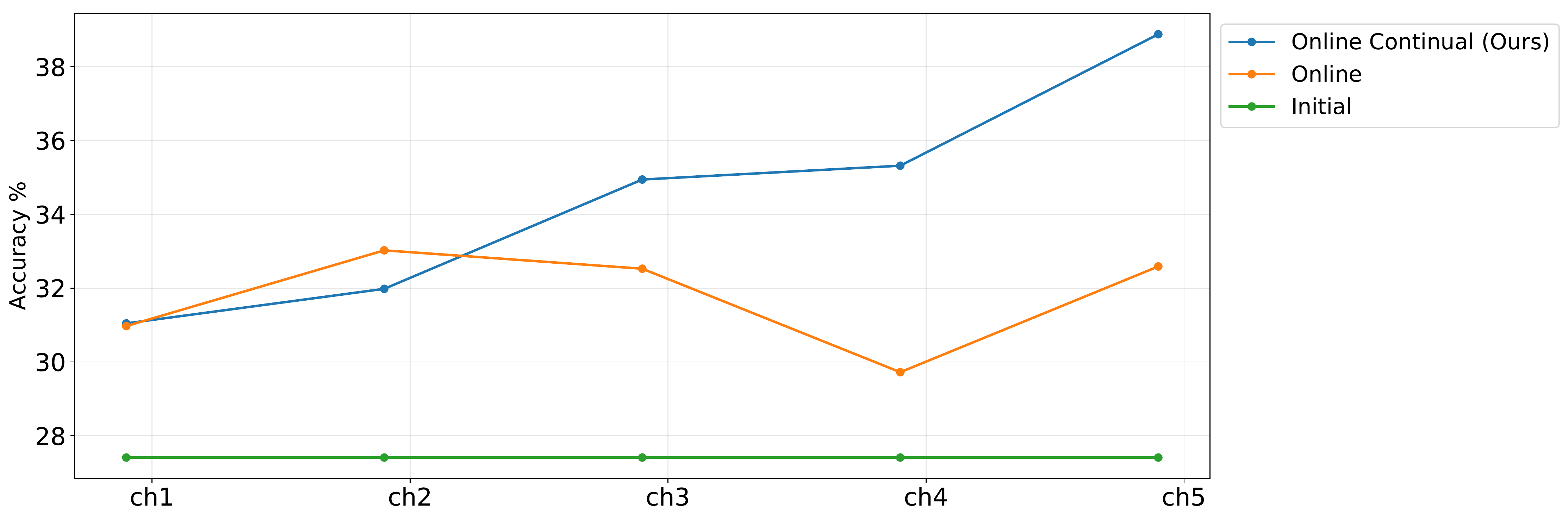} }}%
   \subfloat[]{\includegraphics[width=0.355\textwidth]{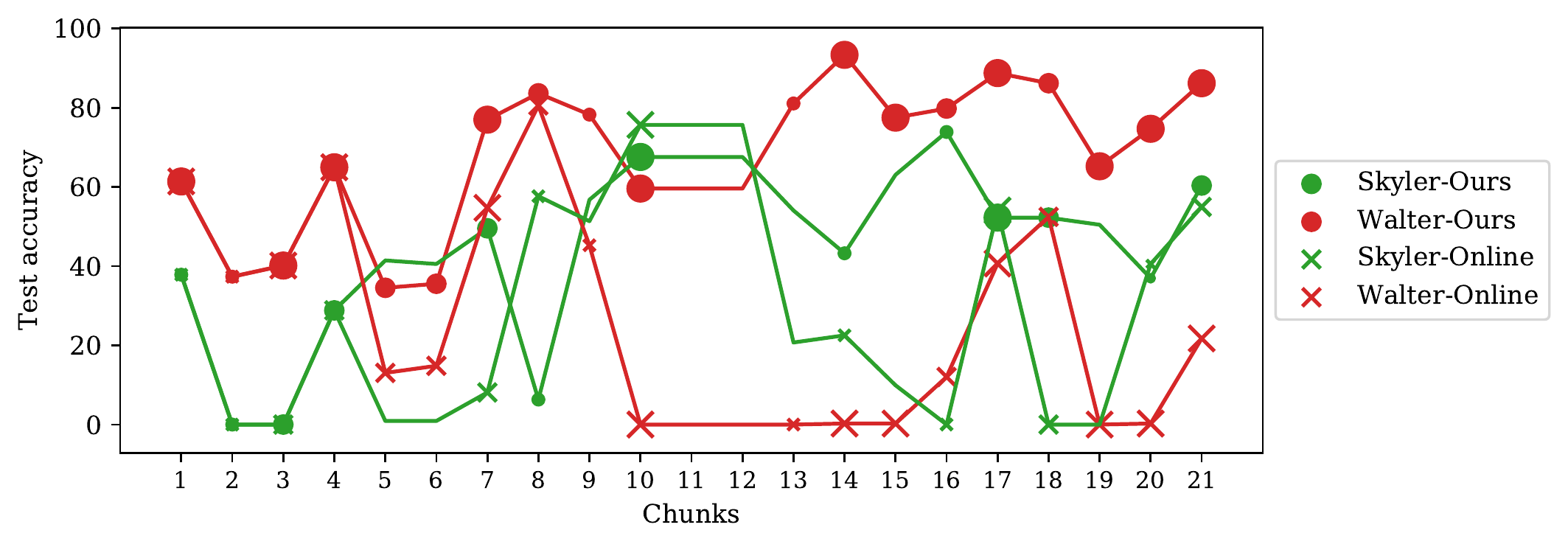}}
   \vspace*{-0.4cm}
    \caption{\footnotesize Self-supervised setting: accuracy on all faces of Big Bang Theory after each episode (a) and of Mad Men after each of the 5 chunks (b). (c) a study on the actors recognition during the course of training, figure shows two main actors test accuracy after each chunk in BB.}
    \label{fig:selfsupervised}%
    \vspace*{-0.4cm}
\end{figure*}
\subsubsection{Weak Supervision Results}
Figure~\ref{fig:soapseries_super} (a) shows the actor recognition accuracy evaluated on all the test data of BBT, at the end of each episode.
Initially, the Online Baseline (orange) 
obtains an increase of $20\%$ in accuracy compared to the initial model. Yet it fails to continue accumulating knowledge and improving the accuracy as training continues.
After the third episode, the overall accuracy starts to decay, probably because the knowledge learned from these new episodes interferes with what was learned previously. In contrast, our Online Continual Learning system (blue) continues to improve its performance and achieves at the end of the 6 episodes an  accuracy that matches the accuracy of the model trained with shuffled data under the i.i.d. condition (Online Joint Training, pink). Offline Joint Training (purple) with multiple revisits to the shuffled data achieves the top performance. Note that this is only $8\%$ higher than our continual learning system trained under the online and changing distribution condition. 

Figure~\ref{fig:soapseries_super} (b) shows the accuracy on all the test data of BB, after each 4 chunks while learning the 4 episodes.
Clearly this tv series is much harder than BBT. Most of the shots are outdoor and under varied lighting conditions, as also noted in~\cite{aljundi2016actor}. This corresponds to large distribution changes within and between episodes.  Here, the Online Baseline (orange) fails to increase the performance after the first episode. Its accuracy notably fluctuates, probably depending on how (un)related the recently seen data is to the rest of the series. Again, our Online Continual Learning system (blue) succeeds in improving and accumulating knowledge -- up to a $100\%$ improvement over the Online Baseline. Like the Online Baseline, its performance drops at times, yet the drops are dampened significantly, allowing the model to keep on learning over time. 
Surprisingly, it even outperforms the Online Joint Training baseline (pink) and comes close to the Offline Joint training upper bound (purple) that only reaches this accuracy after ten revisits to the training data.

\subsubsection{Self Supervision Results}
Next we move to the case with self-supervision.
This scenario reflects the ideal case where continual learning becomes most interesting. 
Remember that, as a clue for self-supervision, we use the fact that multiple tracks appearing in the same image should have different identities. 
We use the six episodes of BBT, although only the first and the sixth episodes actually have a good number of tracks with two persons appearing in one image. 
Figure~\ref{fig:selfsupervised} (a) shows the accuracy on all the episodes after learning each episode. Note how the Online Learning Baseline (orange) continues to improve slightly as more episodes are watched. It's only when we get to the last episode, with a larger number of useful tracks, 
that our Continual Online Learning (blue) starts to outperform the Online Learning Baseline.

Figure~\ref{fig:selfsupervised} (b) shows the recognition accuracy on the first episode of Mad Men after each chunk. Similar to the previous experiments our Online Continual learning (blue) succeeds in improving the performance and accumulating the knowledge. We conclude that the ability of continual learning of stabilizing the online learning is clearly shown, both for weak and self-supervised scenarios. 
\subsubsection{Ablation Study}
Next we perform an ablation study to evaluate the impact of two components of our system. The first factor is the hard buffer used for stabilizing the online training and for updating the importance weights. The second factor is the mechanism for accumulating importance weights across updates. In our system we use a cumulative moving average, which gives all the estimated importance weights the same weight. An alternative is to deploy a decaying average. This reduces the impact of old importance weights in favor of the newest ones. To this end, one can set $\Omega_t=(\Omega_{t-1} +\Omega^*)/2$ where $\Omega^*$ are the currently estimated importance weights. Figure~\ref{fig:soapseries_super}(c) shows the accuracy on all the test data of BBT after each episode achieved by the different variants. The hard buffer clearly improves the performance of both the Online Baseline and Online Continual learning. The buffer with hard samples, even if small, gives the learner a chance to re-pass over hard samples and to adjust its gradients for a better estimate of the parameters update step. Additionally, it allows a better estimate of the importance weights used in our Online Continual Learning.  
The decaying average for the importance weights update, leads to more fluctuations due to the higher impact of more recent importance estimates. 
This allows more forgetting and more bias towards the recent estimate that could be unrepresentative to the overall test data.

\myparagraph{Relationship between samples and recognition performance during training:}
To show how the predictions on the seen actors change over the online training time,  we plotted the accuracies per actor after each chunk (for two most frequent characters of BB, to avoid overloading the figure), see Fig.\ref{fig:selfsupervised}(c). Marker size indicates the actor's frequency in a chunk; no marker indicates zero appearance. Low frequency in a chunk typically causes the accuracy of the online baseline to drop while our method is more stable.

\subsection{Monocular Collision Avoidance}\label{sec:experiments:collisionavoidance}
Collision avoidance is the task in which a robot navigates through the environment aimlessly while avoiding obstacles. We train a neural network to perform this task, at test time, based on single RGB images.
Training is done with self-supervision where a simple heuristic based on extra sensors, serves as a supervising expert.
The deep neural network learns to imitate the expert's control, so cloning its behavior.
The task of collision avoidance is best demonstrated in a variety of environments.
However, hardware or legal constraints might prevent storing all training data, urging the need for an online learning setup. As the network tends to forget what it has learned over time,
the setup is excellent for testing online continual learning.

\begin{figure}
    \centering
    \includegraphics[width=0.45\textwidth]{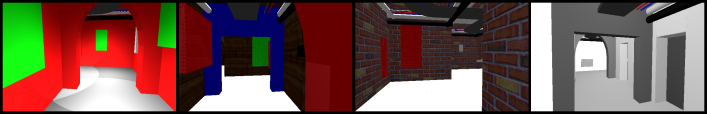}
    \caption{\footnotesize Example views in the corridor sequence corresponding to environments A, B, C and D, depicted from left to right.}
    \label{fig:corridors}
    \vspace*{-0.4cm}
\end{figure}

\begin{figure*}
 \vspace*{-0.4cm}
    \centering
    \includegraphics[width=0.9\textwidth]{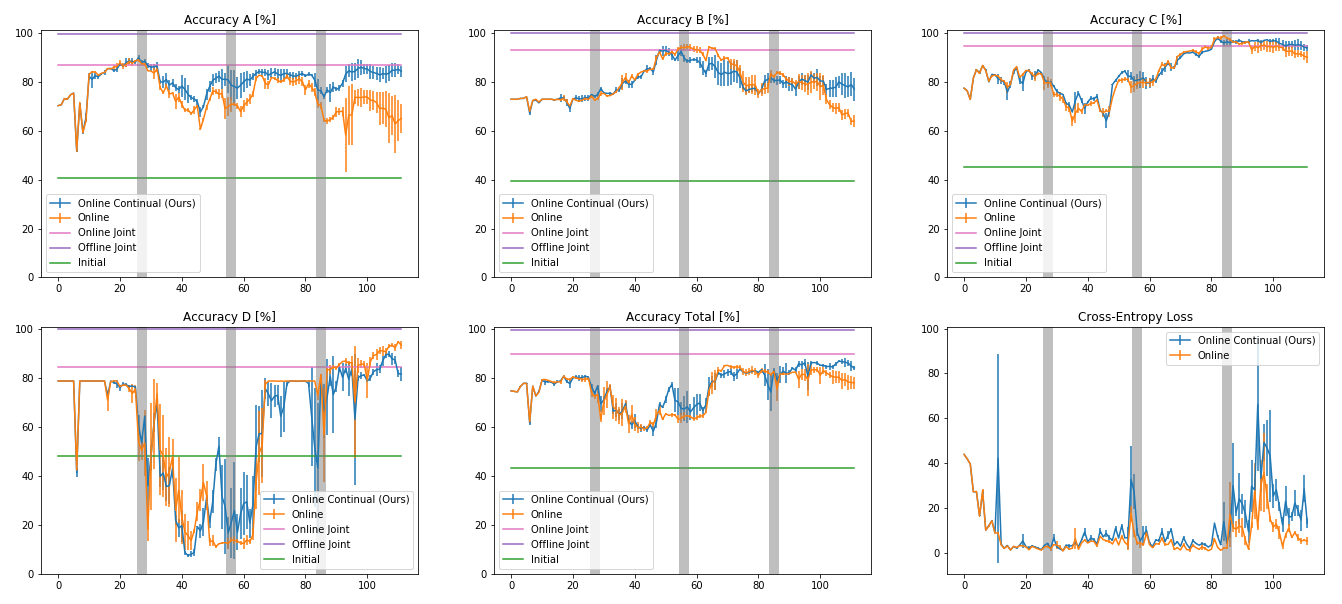}
    \caption{\footnotesize The training accuracies on the different types of corridors as well as the total accuracy during training on the corridor sequence (A,B,C,D) as depicted in figure \ref{fig:corridors}. Grey lines indicate the transition to a new environment. The lower right figure shows the cross-entropy loss on the recent buffer. The accuracies of the baselines are added as horizontal lines for the initial model, the jointly trained model both online and offline.}
    \label{fig:corridoraccuracies}
    \vspace*{-0.4cm}
\end{figure*}

\myparagraph{Architecture:}
Our model takes a 128x128 RGB frame as input and outputs three discrete steering directions.
The architecture consists of 2 convolutional and 2 fully-connected layers with ReLU-activations. 
The training starts with random initialization of the weights and continues with gradient descent on a cross-entropy loss.

\myparagraph{Simulation:}
The experiment is done in a Gazebo simulated environment with the Hector Quadrotor model.
The expert is a heuristic reading scans from a Laser Range Finder mounted on the drone and turning towards the direction with the furthest depth.
The demonstration of the expert follows a sequence of four different corridors, referred to as A,B,C and D. 
The environments differ in texture, obstacles and turns, as visible in figure \ref{fig:corridors}.

\myparagraph{Training:}
Every 10 steps a backward pass occurs, minimizing the cross-entropy loss, shown in the lower right of figure \ref{fig:corridoraccuracies}. For each model, three networks are trained with different seeds resulting in the error bars plotted. 

\myparagraph{Test:}
The models are evaluated on the entire data sequence as reported in figure \ref{fig:corridoraccuracies}.
The grey bars on the x-axis indicate crossings to new environments. Besides the general online with no continual learning baseline, the performance of following models are given: a scratch initialized model, an online jointly trained model as well as offline. The online joint model has seen all the data once but in an i.i.d. manner.
The accuracy of both the online with and without continual learning increases in environments where it is currently learning. Online training without continual training, however, tends to forget the early environments like A, B and C while training in new environments.
Especially in environment B and D, the effect is outspoken.
The cross-entropy loss in environment D rises for all models, indicating a significant change in the data.


\begin{figure}
    \centering
    \includegraphics[width=0.45\textwidth]{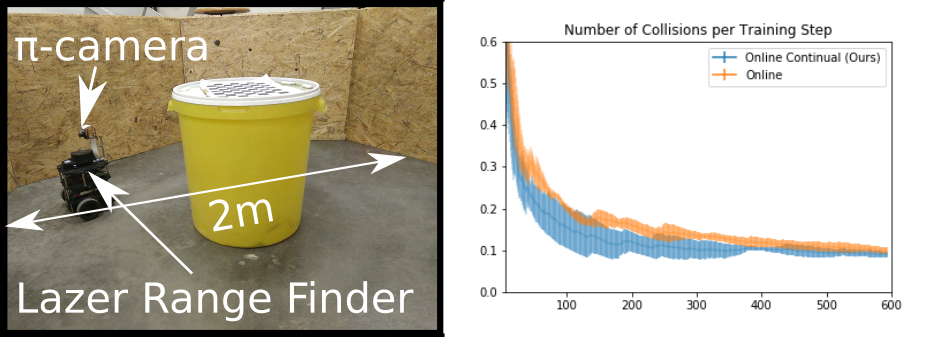}
     \caption{\footnotesize Left: Real-world online and on-policy setup. Right: Number of collisions per training step. Using online continual learning speeds up the training.}
    \label{fig:realworld}
    \vspace*{-0.5cm}
\end{figure}



\subsection{Proof Of Concept in the Real World}\label{sec:experiments:realworld}
In a final experiment, we apply online continual learning on a turtlebot in a small arena in our lab, see figure \ref{fig:realworld}. The model is on-policy pretrained in a similar simulated environment without continual learning. On-policy refers to the model being in control during training instead of the expert.
In the previous experiments, continual learning has proven to be advantageous when big differences occur in the data. In this setup we show that continual learning also provides stabilization during on-policy training within one environment. 
Again, an expert based on the Laser Range Finder is providing a self-supervisory signal.
On-policy learning tends to be more difficult as the data contains a lot of ``dummy'' samples when the model visits irrelevant states.
This data inefficiency causes the model to learn slower and possibly forget along the way.
For example, if the model collides on the left side, the recent data teaches the model to turn right more often.
However, after crossing the arena and bumping on the right side, you still want the model to remember its mistakes made earlier. 
As such, preserving acquired knowledge over time is crucial for on-policy online learning.
In figure \ref{fig:realworld}, we show the number of collisions per step over time with error bars taken over three different models.
Clearly continual learning helps the model to learn faster, with the number of collisions dropping faster than without it.\\[-20pt]
\section{Discussion and Conclusion}\label{sec:discussion}

The importance weight regularization appears most effective in online training scenarios when large changes in the learned distribution occur. The closer the online data stream is to i.i.d. samples, the smaller the positive continual learning effect. 

In some cases however, continual learning tends to slow down the adaptation to newly seen data. Especially when the new data is much more informative or representative than the old, continual learning initially has a negative effect on the training. 
In other words, pure online learning is faster to adapt to new changes but therefore also inherently less stable. Ultimately, whether the stabilizing effect of continual learning is advantageous or not, depends on the time scale of the changes in the data.

While in this work we focus on a setting where the network architecture remains fixed, and no new outputs or tasks are added over time, we believe it could also be applied in other settings. 
For instance in a class-incremental setting, an extra head could be added to the network each time a new category label appears. 
Alternatively, a projection into an embedding space could be used, as in~\cite{Elhoseiny2018Exploring}, avoiding the need for a growing network architecture. These are directions for future work.




Due to the limited time, we used data from published datasets in the face recognition experiment allowing quantitative evaluation. However, as future work, we plan to test self-supervised online continual learning on large scale tv-series, thus learning for a longer time.

In conclusion, we pushed the limits of current task-based sequential learning towards online task-free continual learning. We assume an infinite stream of input data, containing changes in the input distribution both gradual and sudden. 
Our protocol deploys a state of the art importance weight regularization method for online continual learning by detecting when, how and on what data to perform importance weight updates.
Its effectiveness is validated successfully for both supervised and self-supervised learning. More specifically, by using our continual learning method, we demonstrate an improvement of stability and performance over the baseline in applications like learning face identities from watching tv-series and robotic collision avoidance.

\small {\textbf{Acknowledgments:}
Rahaf Aljundi's PhD is funded by an FWO scholarship.
This  work  was further supported  by  the  CAMETRON research project (GOA) of the KU Leuven and the FWO SBO project Omnidrone.}
 
{\small
\bibliographystyle{ieee}
\bibliography{egbib}

\begin{thebibliography}{10}\itemsep=-1pt

\bibitem{aljundi2017memory}
R.~Aljundi, F.~Babiloni, M.~Elhoseiny, M.~Rohrbach, and T.~Tuytelaars.
\newblock Memory aware synapses: Learning what (not) to forget.
\newblock {\em arXiv preprint arXiv:1711.09601}, 2017.

\bibitem{aljundi2016expert}
R.~Aljundi, P.~Chakravarty, and T.~Tuytelaars.
\newblock Expert gate: Lifelong learning with a network of experts.
\newblock In {\em IEEE Conference on Computer Vision and Pattern Recognition
  (CVPR)}, 2016.

\bibitem{aljundi2016actor}
R.~Aljundi, P.~Chakravarty, and T.~Tuytelaars.
\newblock Who’s that actor? automatic labelling of actors in tv series
  starting from imdb images.
\newblock In {\em Asian Conference on Computer Vision}, pages 467--483.
  Springer, 2016.

\bibitem{balntas2016learning}
V.~Balntas, E.~Riba, D.~Ponsa, and K.~Mikolajczyk.
\newblock Learning local feature descriptors with triplets and shallow
  convolutional neural networks.
\newblock {\em BMVC}, pages 119.1--119.11, 01 2016.

\bibitem{bottou1998online}
L.~Bottou.
\newblock Online learning and stochastic approximations.
\newblock {\em On-line learning in neural networks}, 17(9):142, 1998.

\bibitem{duchi2011adaptive}
J.~Duchi, E.~Hazan, and Y.~Singer.
\newblock Adaptive subgradient methods for online learning and stochastic
  optimization.
\newblock {\em Journal of Machine Learning Research}, 12(Jul):2121--2159, 2011.

\bibitem{Elhoseiny2018Exploring}
M.~Elhoseiny, F.~babiloni, R.~Aljundi, M.~Rohrbach, and T.~Tuytelaars.
\newblock Exploring the challenges towards lifelong fact learning.
\newblock In {\em Asian Conference on Computer Vision}, 2018.

\bibitem{french1999catastrophic}
R.~M. French.
\newblock Catastrophic forgetting in connectionist networks.
\newblock {\em Trends in cognitive sciences}, 3(4):128--135, 1999.

\bibitem{hazan2008adaptive}
E.~Hazan, A.~Rakhlin, and P.~L. Bartlett.
\newblock Adaptive online gradient descent.
\newblock In {\em Advances in Neural Information Processing Systems}, pages
  65--72, 2008.

\bibitem{rainbow17hessel}
M.~Hessel, J.~Modayil, H.~van Hasselt, T.~Schaul, G.~Ostrovski, W.~Dabney,
  D.~Horgan, B.~Piot, M.~Azar, and D.~Silver.
\newblock Rainbow: Combining improvements in deep reinforcement learning.
\newblock {\em arXiv preprint arXiv:1710.02298}, 2017.

\bibitem{hsu2018re}
Y.-C. Hsu, Y.-C. Liu, and Z.~Kira.
\newblock Re-evaluating continual learning scenarios: A categorization and case
  for strong baselines.
\newblock {\em arXiv preprint arXiv:1810.12488}, 2018.

\bibitem{hu2011online}
T.~Hu.
\newblock Online regression with varying gaussians and non-identical
  distributions.
\newblock {\em Analysis and Applications}, 9(04):395--408, 2011.

\bibitem{kirkpatrick2016overcoming}
J.~Kirkpatrick, R.~Pascanu, N.~Rabinowitz, J.~Veness, G.~Desjardins, A.~A.
  Rusu, K.~Milan, J.~Quan, T.~Ramalho, A.~Grabska-Barwinska, et~al.
\newblock Overcoming catastrophic forgetting in neural networks.
\newblock {\em arXiv preprint arXiv:1612.00796}, 2016.

\bibitem{kivinen2004online}
J.~Kivinen, A.~J. Smola, and R.~C. Williamson.
\newblock Online learning with kernels.
\newblock {\em IEEE transactions on signal processing}, 52(8):2165--2176, 2004.

\bibitem{krizhevsky2014one}
A.~Krizhevsky.
\newblock One weird trick for parallelizing convolutional neural networks.
\newblock {\em arXiv preprint arXiv:1404.5997}, 2014.

\bibitem{krizhevsky2012imagenet}
A.~Krizhevsky, I.~Sutskever, and G.~E. Hinton.
\newblock Imagenet classification with deep convolutional neural networks.
\newblock In {\em Advances in neural information processing systems}, pages
  1097--1105, 2012.

\bibitem{lee2017overcoming}
S.-W. Lee, J.-H. Kim, J.-W. Ha, and B.-T. Zhang.
\newblock Overcoming catastrophic forgetting by incremental moment matching.
\newblock {\em arXiv preprint arXiv:1703.08475}, 2017.

\bibitem{li2018online}
W.~Li, J.~Huo, Y.~Shi, Y.~Gao, L.~Wang, and J.~Luo.
\newblock Online deep metric learning.
\newblock {\em arXiv preprint arXiv:1805.05510}, 2018.

\bibitem{li2016learning}
Z.~Li and D.~Hoiem.
\newblock Learning without forgetting.
\newblock In {\em European Conference on Computer Vision}, pages 614--629.
  Springer, 2016.

\bibitem{lopez2017gradient}
D.~Lopez-Paz et~al.
\newblock Gradient episodic memory for continual learning.
\newblock In {\em Advances in Neural Information Processing Systems}, pages
  6470--6479, 2017.

\bibitem{mccloskey1989catastrophic}
M.~McCloskey and N.~J. Cohen.
\newblock Catastrophic interference in connectionist networks: The sequential
  learning problem.
\newblock {\em Psychology of learning and motivation}, 24:109--165, 1989.

\bibitem{dqn2014mnih}
V.~Mnih, K.~Kavukcuoglu, D.~Silver, A.~A. Rusu, J.~Veness, M.~G. Bellemare,
  A.~Graves, M.~Riedmiller, A.~K. Fidjeland, G.~Ostrovski, S.~Petersen,
  C.~Beattie, A.~Sadik, I.~Antonoglou, H.~King, D.~Kumaran, D.~Wierstra,
  S.~Legg, and D.~Hassabis.
\newblock Human-level control through deep reinforcement learning.
\newblock {\em Nature}, 2014.

\bibitem{Moore1993}
A.~W. Moore and C.~G. Atkeson.
\newblock Prioritized sweeping: Reinforcement learning with less data and less
  time.
\newblock {\em Machine Learning}, 13(1):103--130, Oct 1993.

\bibitem{pentina15nips}
A.~Pentina and C.~H. Lampert.
\newblock Lifelong learning with non-i.i.d. tasks.
\newblock {\em NIPS}, 2015.

\bibitem{pernici2018memory}
F.~Pernici, F.~Bartoli, M.~Bruni, and A.~D. Bimbo.
\newblock Memory based online learning of deep representations from video
  streams.
\newblock {\em CoRR}, abs/1711.07368, 2017.

\bibitem{pernici17video}
F.~Pernici and A.~Del~Bimbo.
\newblock Unsupervised incremental learning of deep descriptors from video
  streams.
\newblock {\em ICMEW.2017.8026276.}, pages 477--482, 2017.

\bibitem{ramasamy2018online}
S.~Ramasamy, K.~Rajaraman, P.~Krishnaswamy, and V.~Chandrasekhar.
\newblock Online deep learning: growing rbm on the fly.
\newblock {\em arXiv preprint arXiv:1803.02043}, 2018.

\bibitem{rannen2017encoder}
A.~Rannen, R.~Aljundi, M.~B. Blaschko, and T.~Tuytelaars.
\newblock Encoder based lifelong learning.
\newblock In {\em Proceedings of the IEEE Conference on Computer Vision and
  Pattern Recognition}, pages 1320--1328, 2017.

\bibitem{rebuffi2016icarl}
S.-A. Rebuffi, A.~Kolesnikov, and C.~H. Lampert.
\newblock icarl: Incremental classifier and representation learning.
\newblock {\em arXiv preprint arXiv:1611.07725}, 2016.

\bibitem{robins1995catastrophic}
A.~Robins.
\newblock Catastrophic forgetting, rehearsal and pseudorehearsal.
\newblock {\em Connection Science}, 7(2):123--146, 1995.

\bibitem{rolnick17continualexpreplay}
D.~Rolnick, A.~Ahuja, J.~Schwarz, T.~P. Lillicrap, and G.~Wayne.
\newblock Unsupervisedexperience replay for continual learning.
\newblock {\em arxiv:1811.11682.}, 2018.

\bibitem{sahoo2017online}
D.~Sahoo, Q.~Pham, J.~Lu, and S.~C. Hoi.
\newblock Online deep learning: Learning deep neural networks on the fly.
\newblock {\em arXiv preprint arXiv:1711.03705}, 2017.

\bibitem{shalev2012online}
S.~Shalev-Shwartz et~al.
\newblock Online learning and online convex optimization.
\newblock {\em Foundations and Trends{\textregistered} in Machine Learning},
  4(2):107--194, 2012.

\bibitem{silver2013lifelong}
D.~L. Silver, Q.~Yang, and L.~Li.
\newblock Lifelong machine learning systems: Beyond learning algorithms.
\newblock In {\em AAAI Spring Symposium: Lifelong Machine Learning}, pages
  49--55. Citeseer, 2013.

\bibitem{sprechmann2018memory}
P.~Sprechmann, S.~M. Jayakumar, J.~W. Rae, A.~Pritzel, A.~P. Badia, B.~Uria,
  O.~Vinyals, D.~Hassabis, R.~Pascanu, and C.~Blundell.
\newblock Memory-based parameter adaptation.
\newblock {\em arXiv preprint arXiv:1802.10542}, 2018.

\bibitem{strehl2008online}
A.~L. Strehl and M.~L. Littman.
\newblock Online linear regression and its application to model-based
  reinforcement learning.
\newblock In {\em Advances in Neural Information Processing Systems}, pages
  1417--1424, 2008.

\bibitem{thrun1995lifelong}
S.~Thrun and T.~M. Mitchell.
\newblock Lifelong robot learning.
\newblock {\em Robotics and autonomous systems}, 15(1-2):25--46, 1995.

\bibitem{yoon2018lifelong}
J.~Yoon, E.~Yang, J.~Lee, and S.~J. Hwang.
\newblock Lifelong learning with dynamically expandable networks.
\newblock {\em arXiv preprint arXiv:1708.01547}, 2018.

\bibitem{zenke2017improved}
F.~Zenke, B.~Poole, and S.~Ganguli.
\newblock Improved multitask learning through synaptic intelligence.
\newblock In {\em Proceedings of the International Conference on Machine
  Learning (ICML)}, 2017.

\end{thebibliography}
}

\newpage
\newpage\phantom{blabla}

\title{Supplementary Materials}
\author{}
\date{}
\maketitle

\begin{table}[b]
    \centering
    \begin{tabular}{|r|l|l|l|}
         \hline
        & Exp 1 & Exp 2 & Exp 3 \\
         \hline
         Architecture & Alexnet & Tiny v2 & Tiny v2 \\
         Initialization & imagenet & random & random \\
         Learning rate & 0.0001 & 0.01 & 0.01 \\
         Optimizer & SGD & SGD & SGD \\
         Hard Buffer Size & 100 & 40 & 30 \\
         Regularization Weight & 100 & 0.5 & 0.5 \\
         Threshold Mean Loss & 0.3 & 0.5 & 0.5 \\
         Threshold Variance Loss & 0.1 & 0.1 & 0.02 \\
         Length Window Loss & 5 & 5 & 5 \\
         \hline
    \end{tabular}
    \caption{Hyperparameters for different experiments: exp 1 $\sim$ Soap Series (4.1), exp 2 $\sim$ Simulated Corridor (4.2) and exp 3 $\sim$ Real Turtlebot (4.3).}
    \label{tab:hyperparamsimul}
\end{table}

These supplementary materials contain the following extra information: 
\begin{itemize}
    \item Hyperparameters and architectural details for the experiments from section 4.
    \item Example images of the soap series.
    \item Results on collision avoidance in an extra lengthy simulated corridor.
    \item Details and extra results on the real-world collision avoidance with the Turtlebot.
    \item Closing discussion and guidelines on the application of continual learning in an online setting.
\end{itemize}

\section{Hyper-parameters and architectural details}

As to be able to reproduce the results, we provide the reader with the used hyperparameters and networks, see table \ref{tab:hyperparamsimul}.
Regularization weight corresponds to $\lambda$, the continual learning weight in Equation 5. 

\begin{figure*}[h]
    \centering
    \includegraphics[width=\textwidth]{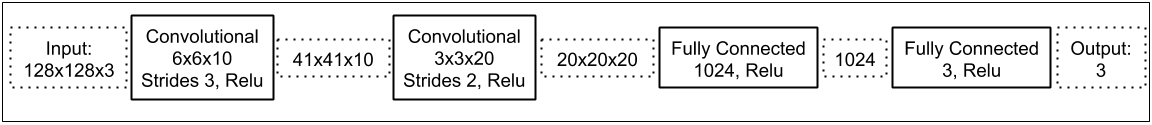}
    \caption{\footnotesize Architecture of the Tiny v2 network used in the monocular collision avoidance experiments (4.2 and 4.3).}
    \label{fig:tinyv2}
\end{figure*}

The Tiny v2 network for the collision avoidance task is a network build especially small in order to allow faster training. The details of the network can be found in figure \ref{fig:tinyv2}.

\section{Examples of the soap series data (Sec.~4.1)}
\begin{figure*}[h]
    \centering
    \includegraphics[width=\textwidth]{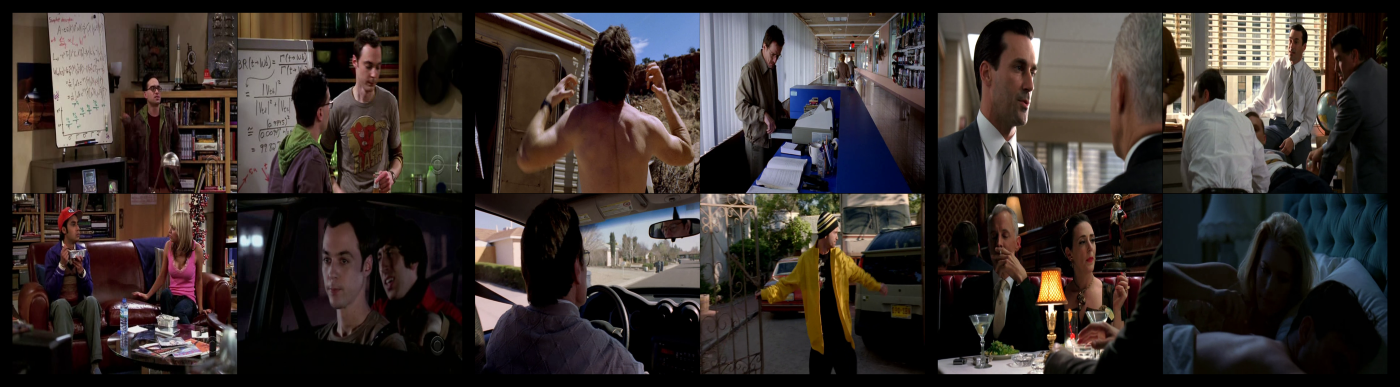}
    \caption{\footnotesize Four example images for each soap series, from left to right: Big Bang Theory, Breaking Bad and Mad Men.}
    \label{fig:soapseries}
\end{figure*}

In figure \ref{fig:soapseries}, 4 example frames are shown for each of the different soap series: Big Bang Theory, Breaking Bad and Mad Men. 
These examples demonstrate the scene diversity and large variance in imaging conditions. As mentioned in the paper, Breaking Bad is more actor-centric with a majority of the frames showing only the main character, making it less suited for the self-supervised setup.

\section{Larger experiment on collision avoidance in simulation (Sec.~4.2)}

To demonstrate both the strengths and weaknesses of our continual learning method, we expanded the corridor experiment of Sec.~4.2 to a sequence of 10 corridors, equal to about 20 minutes flying time or around 10.000 frames. The sequence of different corridors is depicted in figure \ref{fig:corridorsexamples}, exhibiting a large variety in textures and obstacles. The length of the sequence allows us to see the longer trend of continual learning.

While training the models online, we evaluate the accuracy on different corridors separately. 
Due to an imbalance over actions within one corridor, we perform an evaluation based on the total accuracy averaged over the different actions, referred as 'Weighted Accuracy'. When a model becomes degenerated, thus only predicting the most common action in a corridor, an unnormalized accuracy would remain high.

We observe that this data imbalance also affects the online learning as often multiple gradient steps are taken in favor of only some actions. To bypass this impediment, we experiment here with an additional normalization constraint on the hard buffer forcing an equal distribution over all actions. 

\myparagraph{Results}
\begin{figure}[h]
    \centering
    \includegraphics[width=0.35\textwidth]{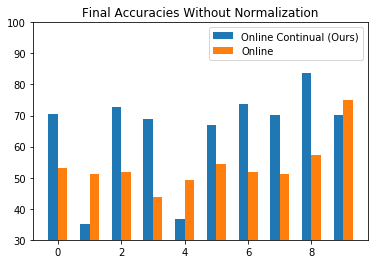}
    \includegraphics[width=0.35\textwidth]{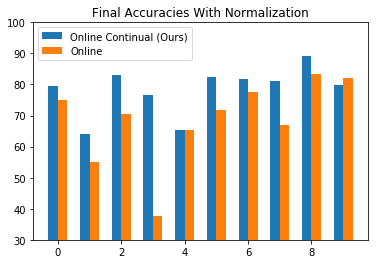}
    \caption{\footnotesize Accuracy's for all 10 corridors at the end of training on the corridor sequence without (top) and with (bottom) the normalization constraint on the hard buffer.}
    \label{fig:finalaccuracies}
\end{figure}

In fig. \ref{fig:finalaccuracies}, we show the improvement obtained by our proposed online continual learning method over the online baseline, both with and without the normalization constraint on the buffer with hard examples. The bars express the final accuracy of each corridor as a mean over three models trained with different seeds. The normalization constraint has a positive effect on both continual and normal online learning. Our online continual learning process clearly outperforms the online baseline for most of the corridors. Without the action normalization, the models fail to learn certain corridors, like 1 and 4, resulting in no knowledge that can be preserved by our continual learning method. However if the model grasps information while going through a corridor, it succeeds at preserving it with continual learning, outperforming the online baseline with 15 to 20$\%$ accuracy. 
Moreover with the action normalization, continual learning succeeds at acquiring knowledge in each corridor, outperforming the baseline in all but last corridors.

Figure~\ref{fig:corridorsaccuracies} provides a more in-depth analysis. Here, we show the evolution over time of the cross-entropy loss and the total accuracy over all corridors. We also report the evolution over time of the weighted accuracy, for each corridor separately.
From these plots, one can conclude that the buffer normalization clearly has beneficial effects for online learning, especially in the green areas (corresponding to learning taking place on imbalanced corridors).  However, the constraint leaves less room in the hard buffer for recent samples causing a slower adaptation of the model during training, as can be observed in the red areas, allowing the models without normalization to improve faster.

In multiple examples, highlighted with blue, the continual learning allows a preservation of knowledge seen before, demonstrating the success of our method. The trend is most clear for the early corridors as the forgetting tends to be worse over time. This phenomenon is also responsible for the total accuracy reaching 80$\%$ for continual learning instead of only 70$\%$ for normal online learning. This positive trend can be expected to increase when learning over even longer sequences.

In some cases, highlighted in orange, the baseline performance of an old corridor improves while training in a new corridor, reaching a similar accuracy as our continual learning method. In other words, the impact of forgetting seems less as the baseline is able to learn the same knowledge again.

This lengthy experiment demonstrates the strengths of continual learning, including the expected positive trend when applying it to longer sequences of data.

 \section{Collision avoidance on real Turtlebot (Sec.~4.3)}
\begin{figure}[h]
    \centering
    \includegraphics[width=0.45\textwidth]{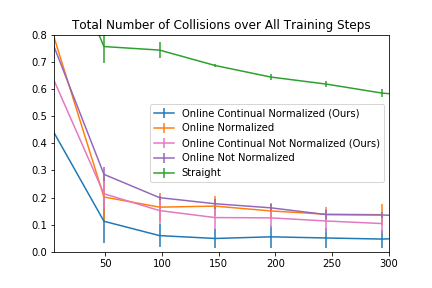}
    \caption{\footnotesize Performance expressed as the average number of collisions - i.e. the total number of collisions divided by the total number of gradient steps.}
    \label{fig:realworld2}
\end{figure}

In this proof-of-concept, an neural network steers a turtlebot around one big yellow object (see figure 7 in the paper). Each frame is kept in a buffer containing the 40 most recent frames combined with expert labels. Every 10 frames a gradient step is taken. When a collision is detected by the Lazer Range Finder, the training is paused and the Turtlebot turns automatically such that the closest obstacle is at its back. 
The hyperparameters can be found in table \ref{tab:hyperparamsimul}. Each model is trained three times and takes about 20 minutes, or 300 gradient steps, till convergence.

Extra results and baselines are shown in figure \ref{fig:realworld2}, plotting the total number of collisions divided by the total number of gradient steps. Driving straight leads to an average of 0.6 collisions per gradient step. Adding action normalization in the hard buffer and applying continual learning both have a clear positive influence. The action normalization allows an even larger improvement of our continual learning method over the baseline. 

This real-world experiment differs in two significant ways from the previous experiments. First, the agent stays in one domain that does not vary over time. Second, the agent acts within the environment to create new data making the setup on-policy and online. Although there are no domain or task changes over time,
our continual learning method has a clear positive effect. This result fully supports our claim of "Task-Free" continual learning, namely that it is not required to have significant changes in your data in order to do better than a normal online learner. The continual learning method inherently stabilizes the online learning in an on-policy setup.

However, a major challenge in online/on-policy learning is dealing with uninformative states. These states lead to less information in a batch and thus slower training. Samples that do contain relevant information, are better preserved in the online-continual setting, resulting in faster learning. Unfortunately, the exact moment of large information gain varies over different runs resulting in a higher variance. This explains the larger variance in figure \ref{fig:realworld2} and \ref{fig:realworld}.

\begin{figure*}[h]
    \centering
    \includegraphics[width=\textwidth]{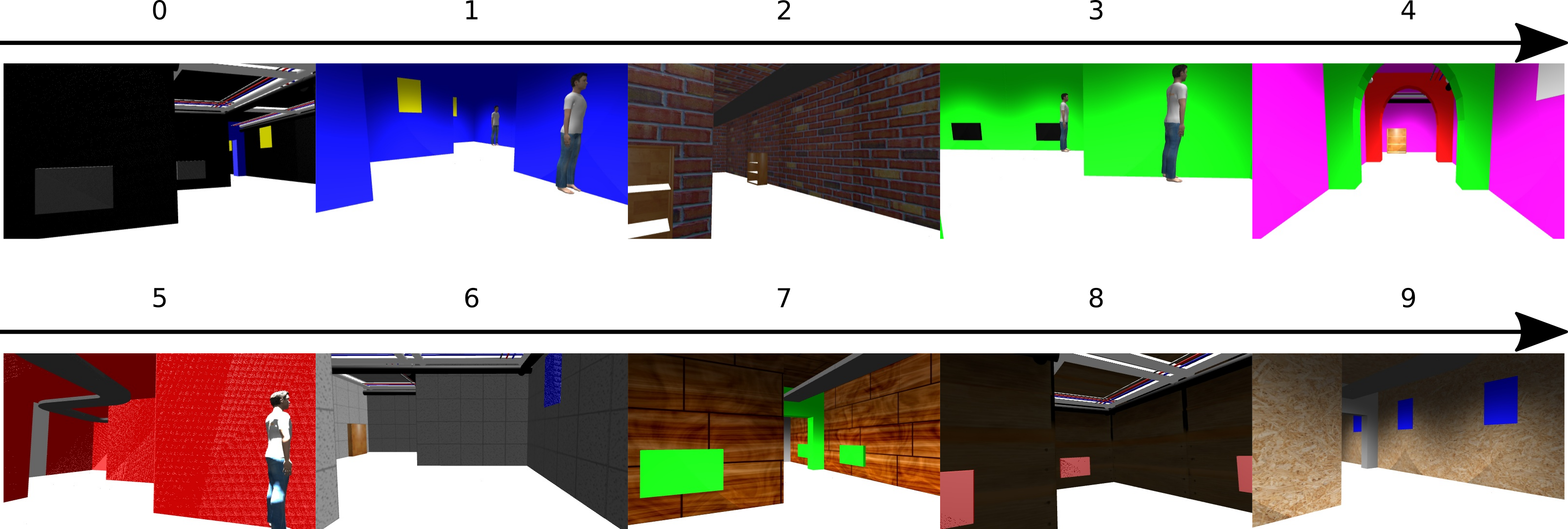}
    \caption{\footnotesize Example views in the longer corridor sequence, corresponding to 10 environments depicted in lexicographic order.}
    \label{fig:corridorsexamples}
    \vspace*{-0.4cm}
\end{figure*}

\begin{figure*}[h]
    \centering
    \includegraphics[width=\textwidth]{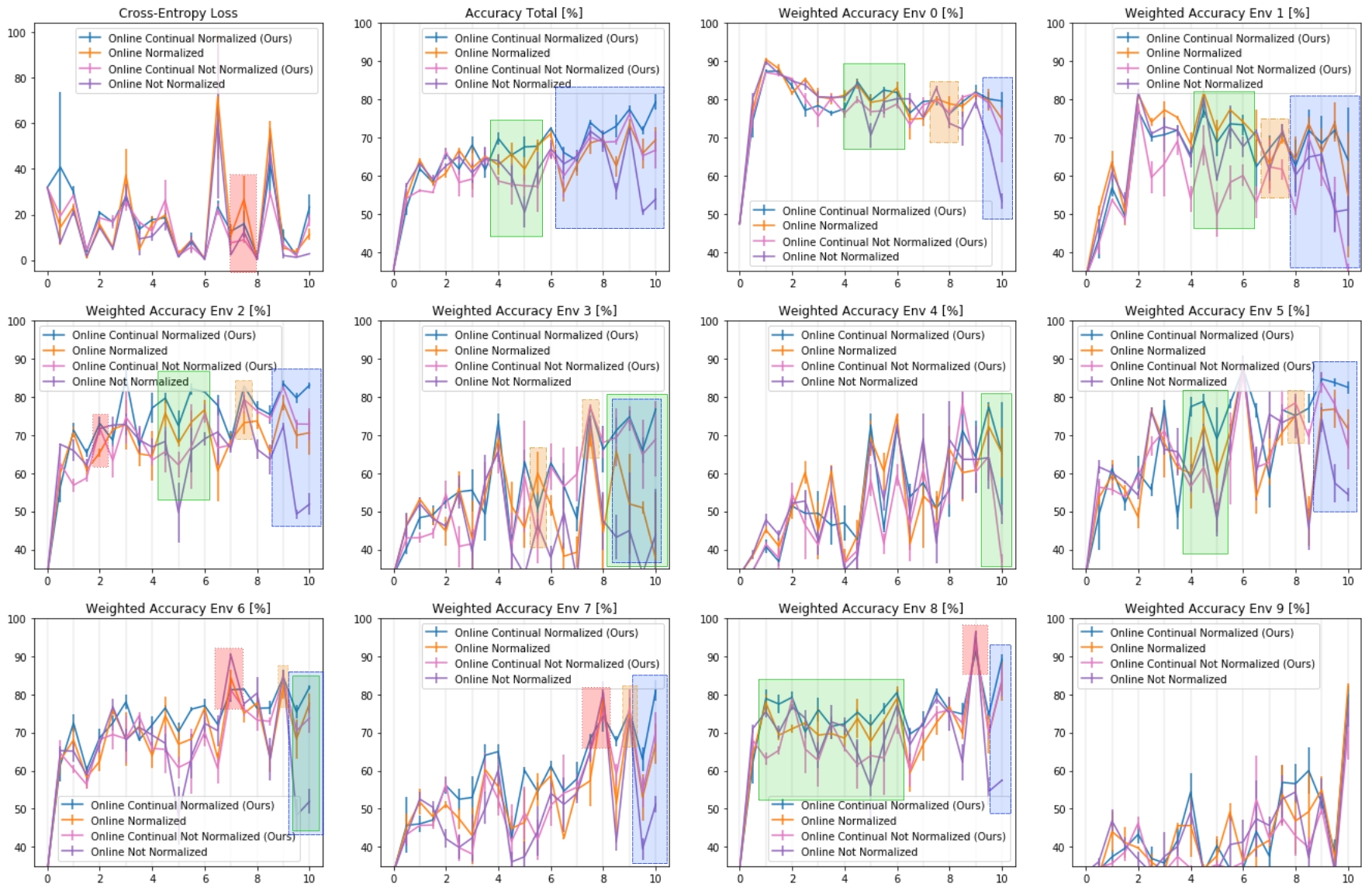}
    \caption{\footnotesize Cross-Entropy loss and accuracy's on total and separate corridors while training online on the sequence of 10 corridors. Blue squares indicate continual learning outperforming baseline models. Green squares indicate positive normalization effects for both continual and baseline models. Red squares indicate slower learning due to normalization constraint. Orange squares indicate learning forgotten knowledge by the baseline model.}
    \label{fig:corridorsaccuracies}
    \vspace*{-0.4cm}
\end{figure*}

\section{Closing discussion / General guidelines}
When considering applying continual learning to a specific problem, it is best to keep two guidelines in mind:

The \textit{mean and variance threshold} of the loss window should be carefully chosen. If both thresholds are too low, the model will not use the MAS regularization; conversely, if too high, the model will slow down the learning by preserving irrelevant information as the importance weights are updated too frequently. 
The latter case in combination with global averaging usually deteriorates the final performance. Therefore, it is recommended to place the threshold low enough while still allowing importance weight updates. As relaxing the threshold, results in more updates, a decaying update rule allows the model to forget irrelevant previous knowledge. In practice, we discovered that the mean threshold could remain quite high, as long as the variance threshold is low.
Moreover, meta-learning techniques, such as learning-to-learn, could automate these settings.

A \textit{trainable task} is a necessary condition: In order to have the MAS regularization exceed in performance over the baseline, the task must be actually trainable. Although this seems obvious, it is far from trivial in an online learning setting, due to the non-i.i.d. nature. Predicting whether continual learning will perform better than typical online learning depends on stable training. 
For instance, in the collision avoidance task, including an action normalization constraint in the hard buffer, clearly improves the stability of online learning.

In conclusion, we successfully extended continual learning to a task-free online learning algorithm and demonstrated its advantage in following applications:
face recognition in soap series, and monocular collision avoidance both on a drone in simulation and on a Turtlebot in the real-world.

\end{document}